PROCEEDINGS    Open Access

# Collapsing ROC approach for risk prediction research on both common and rare variants

Changshuai Wei, Qing Lu*

## Abstract

Risk prediction that capitalizes on emerging genetic findings holds great promise for improving public health and clinical care. However, recent risk prediction research has shown that predictive tests formed on existing common genetic loci, including those from genome-wide association studies, have lacked sufficient accuracy for clinical use. Because most rare variants on the genome have not yet been studied for their role in risk prediction, future disease prediction discoveries should shift toward a more comprehensive risk prediction strategy that takes into account both common and rare variants. We are proposing a collapsing receiver operating characteristic (CROC) approach for risk prediction research on both common and rare variants. The new approach is an extension of a previously developed forward ROC (FROC) approach, with additional procedures for handling rare variants. The approach was evaluated through the use of 533 single-nucleotide polymorphisms (SNPs) in 37 candidate genes from the Genetic Analysis Workshop 17 mini-exome data set. We found that a prediction model built on all SNPs gained more accuracy (AUC = 0.605) than one built on common variants alone (AUC = 0.585). We further evaluated the performance of two approaches by gradually reducing the number of common variants in the analysis. We found that the CROC method attained more accuracy than the FROC method when the number of common variants in the data decreased. In an extreme scenario, when there are only rare variants in the data, the CROC reached an AUC value of 0.603, whereas the FROC had an AUC value of 0.524.

## Background

The completion of hundreds of genome-wide association studies has brought numerous novel disease susceptibility loci to light. Yet for many diseases the common variants that have been identified explain only a small proportion of disease heritability. Additional genetic variants, including rare variants and gene-gene or gene-environment interactions, remain uncovered. Among these, great attention has been given to the rare variants. Current genome-wide association studies include only single-nucleotide polymorphisms (SNPs) with a minor allele frequency (MAF) greater than 5% [1,2]. Within the next few years, whole-genome sequencing will produce millions of rare variants, with the expectation that some of them might explain part of the missing heritability. In fact, experimental studies have already shown that rare variants are associated with complex diseases, such as obesity [3], schizophrenia [4], and colorectal cancer [5].

The uncovered rare variants, particularly those that are yet to be identified by future whole-genome sequencing studies, can be combined with the known common variants and clinical risk factors for more accurate disease prediction. However, few approaches are available for assessing the combined effect of both common and rare variants in early disease prediction. Statistical approaches, such as the collapsing approach [6] and the weighting approach [7], have recently been proposed to assess the association of rare variants with disease.

The collapsing approach first combines all rare variants into a single common variant and then analyzes it with other common variants using multivariate test statistics. Although this approach was originally proposed for genetic association studies, the idea can be used for genetic risk prediction research as well. We here

* Correspondence: qlu@epi.msu.edu
Department of Epidemiology, Michigan State University, East Lansing, MI 48824, USA

develop a collapsing receiver operating characteristic (CROC) approach for risk prediction research that considers both common variants and rare variants. The new approach is an extension of the previously proposed forward receiver operating characteristic (FROC) approach, which was developed using optimal features of the likelihood ratio rule [8,9]. A multistage collapsing procedure is added to the FROC approach to facilitate its use in sequencing data composed of both common and rare variants.

## Methods
### FROC approach

The receiver operating characteristic (ROC) curve is commonly used in genetic risk prediction research to evaluate the accuracy of a risk prediction model. The ROC curve plots the sensitivity of a prediction model against its specificity by continuously changing the cut-off points over the whole range of possible outcomes. When the ROC curve is formed on the likelihood ratio (LR), which is defined as the ratio of the frequency of a particular test outcome in case subjects to that in control subjects, it attains the maximum performance at each cutoff point. The corresponding one-dimensional summary accuracy index, the area under the ROC curve (AUC), is also the highest among that of all approaches [8]. Based on the optimal properties of the LR, we had previously developed a FROC approach for risk prediction on a large number of common genetic variants [9].

The FROC approach uses a computationally efficient algorithm, the forward selection algorithm, to search a large number of genetic predictors for an optimal risk prediction model. The forward selection algorithm starts with a null model and gradually adds new predictors to the model to improve its accuracy. In each step, the forward selection algorithm searches all loci for a locus that most significantly improves the prediction model. The whole selection process continues until adding a new predictor no longer increases the prediction accuracy. A series of prediction models, with different levels of model complexity, are thus obtained. Among these models, the first model with only one predictor is the simplest model, and it has the lowest AUC value; the last model, which has the largest number of predictors, is the most complex model, and it has the highest AUC value. Although the more complex models tend to have an estimated classification accuracy that is higher, they are more likely to overfit the data when there are more risk groups. We therefore used 10-fold cross-validation to identify the best model with the appropriate complexity, and we used that as our final model. Assuming that the best model selects $K$ loci, which consist of $M_k$ multilocus genotypes, we calculate the likelihood ratio $LR_l^k$ for each multilocus genotype $G_l^k$ using:

$$LR_l^k = \frac{P\left(G_l^k \mid D\right)}{P\left(G_l^k \mid \bar{D}\right)}, \qquad 1 \le l \le M_k, \qquad (1)$$

where $D$ and $\bar{D}$ denote disease status and nondisease status, respectively. Each individual who carries a particular type of $M_k$ multilocus genotype is assigned an LR value. Based on the ranks of the LR values, we can form an optimal ROC curve and calculate its corresponding AUC value:

$$AUC_k = \frac{1}{N_D N_{\bar{D}}} \sum_{i=1}^{N_D} \sum_{j=1}^{N_{\bar{D}}} \psi\left(LR_i^k, LR_j^k\right), \qquad (2)$$

where $N_D$ and $N_{\bar{D}}$ are the number of case subjects and control subjects, respectively. The kernel function $\psi$ is given by:

$$\psi\left(LR_i^k, LR_j^k\right) = \begin{cases} 0 & \text{if } LR_i^k < LR_j^k, \\ 0.5 & \text{if } LR_i^k = LR_j^k, \\ 1 & \text{if } LR_i^k > LR_j^k. \end{cases} \qquad (3)$$

### CROC approach

The FROC approach was proposed for risk prediction research on common variants. When both common and rare variants exist, the FROC approach hardly selects rare variants because of their low frequency, which could lead to low accuracy of the prediction model. To deal with both common and rare variants, we have extended the FROC approach and are introducing the CROC approach here. The CROC adopts a multistage collapsing procedure to collapse rare variants into pseudo-common variants and then uses the forward selection algorithm of the FROC approach to search both common and pseudo-common variants for the best prediction model.

The development of the multistage collapsing procedure is based on the ideas of Li and Leal [6]. Similar to the forward selection algorithm, the collapsing procedure selects rare variants in a stepwise manner and then collapses them into a pseudo-common variant. To illustrate the method, we assume that $(k-1)$ rare variants are selected in the previous $(k-1)$ steps and then collapsed into a pseudo-common variant. The pseudo-common variant is defined by:

$$I_i = \begin{cases} 1, & \text{individual } i \text{ carries at least one of the } k-1 \text{ rare variant}, \\ 0, & \text{otherwise}. \end{cases} \qquad (4)$$

The accuracy of the pseudo-common variant is then measured by its AUC value. In step $k$, we search the remaining rare variants for one locus that increases the AUC value most significantly and collapse it into the pseudo-common variant. The procedure keeps collapsing new rare variants into the pseudo-common variant until the AUC value stops increasing. A pseudo-common variant is thus formed.

We repeat the collapsing procedure on the remaining rare variants and generate a set of pseudo-common variants. The multistage collapsing procedure stops when there are no rare variants left in the data. One of the advantages of using a multistage collapsing procedure instead of the original collapsing procedure is that it could potentially consider bidirectional effects. The forward selection algorithm is then used to search both common variants and pseudo-common variants for an optimal risk prediction model. Because the pseudo-common variants have a higher frequency, they are more likely to be selected by the forward selection algorithm, which could result in increased accuracy of the prediction model.

## Results

We evaluated the performance of the CROC approach using the simulated Genetic Analysis Workshop 17 (GAW17) mini-exome sequencing data. The data are composed of 697 individuals, in which 209 individuals make up the case group. Thirty-seven candidate genes were selected based on the simulation results provided. There are 533 SNPs in the candidate genes, including both disease susceptibility and noise loci. The MAFs of these SNPs range from 0.00072 to 0.45122. Among those, 400 SNPs are rare variants (MAF < 0.01). Using the GAW17 data, we investigated whether the accuracy of the risk prediction model could be improved by considering rare variants in the analysis. An additional analysis was also conducted to compare the performance of the CROC and FROC approaches.

### Risk prediction considering rare variants

We started the analysis by forming a risk prediction model on all common variants. To assess the accuracy improvement by adding rare variants, we also built a risk prediction model using all genetic variants. In all, 200 replicates were used for the analysis. Risk prediction models were built based on the first 100 replicates, using both the CROC and FROC approaches; they were then evaluated on the remaining 100 replicates. The reason for evaluating the model on a separate replicate is to ensure that the rare variants are evaluated. Because of the small sample size of the data set, rare variants are commonly carried by one individual or by a small number of individuals. If we split the data into training and testing data sets, then rare variants present in the training data set will likely be absent in the testing data set and therefore cannot be validated. Using two replicates ensures the presence of the same rare variants in both data sets. However, we should note that the estimation might be biased because of the potential correlation between two replicates.

The accuracy of the risk prediction models was averaged over 100 replicates and is summarized in Table 1. The results show that the prediction model built on both common and rare variants using the CROC approach (AUC = 0.605) attains higher accuracy than the one built on common variants alone using either the CROC or the FROC approach (AUC = 0.585). The CROC approach is equivalent to the FROC approach when only common variants are considered. However, the CROC approach outperforms the FROC approach when both common and rare variants are considered (Table 1). With additional rare variants, the risk prediction model built by the FROC approach maintains the same level of accuracy (AUC = 0.585), whereas the model built using the CROC approach has improved accuracy (AUC = 0.605). In addition, the CROC approach requires less computation time (1058 s) than the FROC approach does (1911 s).

### Comparison of CROC and FROC approaches

From Table 1, we also see that the performance of the two approaches is dependent on how many common or rare variants are involved in the analysis. The FROC and CROC approaches are comparable when only common variants are analyzed. However, the CROC approach outperforms the FROC approach when both common and rare variants are considered in the analysis. To further investigate the performance of the two approaches, we gradually dropped out the common variants while keeping all the rare variants in the analysis. We applied the two approaches on subsets with a reduced number of common variants and obtained the averaged AUC values and their 95% confidence intervals. The results are summarized in Figure 1. We found that the CROC approach outperformed the FROC approach in all cases, especially with the lower number

**Table 1 Accuracy improvement in the CROC approach by adding rare variants**

|  | Common SNPs only | | All SNPs | |
| --- | --- | --- | --- | --- |
|  | FROC | CROC | FROC | CROC |
| Mean of AUC value | 0.585 | 0.585 | 0.585 | 0.605 |
| SD of AUC value | 0.048 | 0.048 | 0.048 | 0.023 |
| Running time (s) | 821 | 821 | 1,911 | 1,058 |

The CROC approach is equivalent to the FROC approach when only common variants are considered.

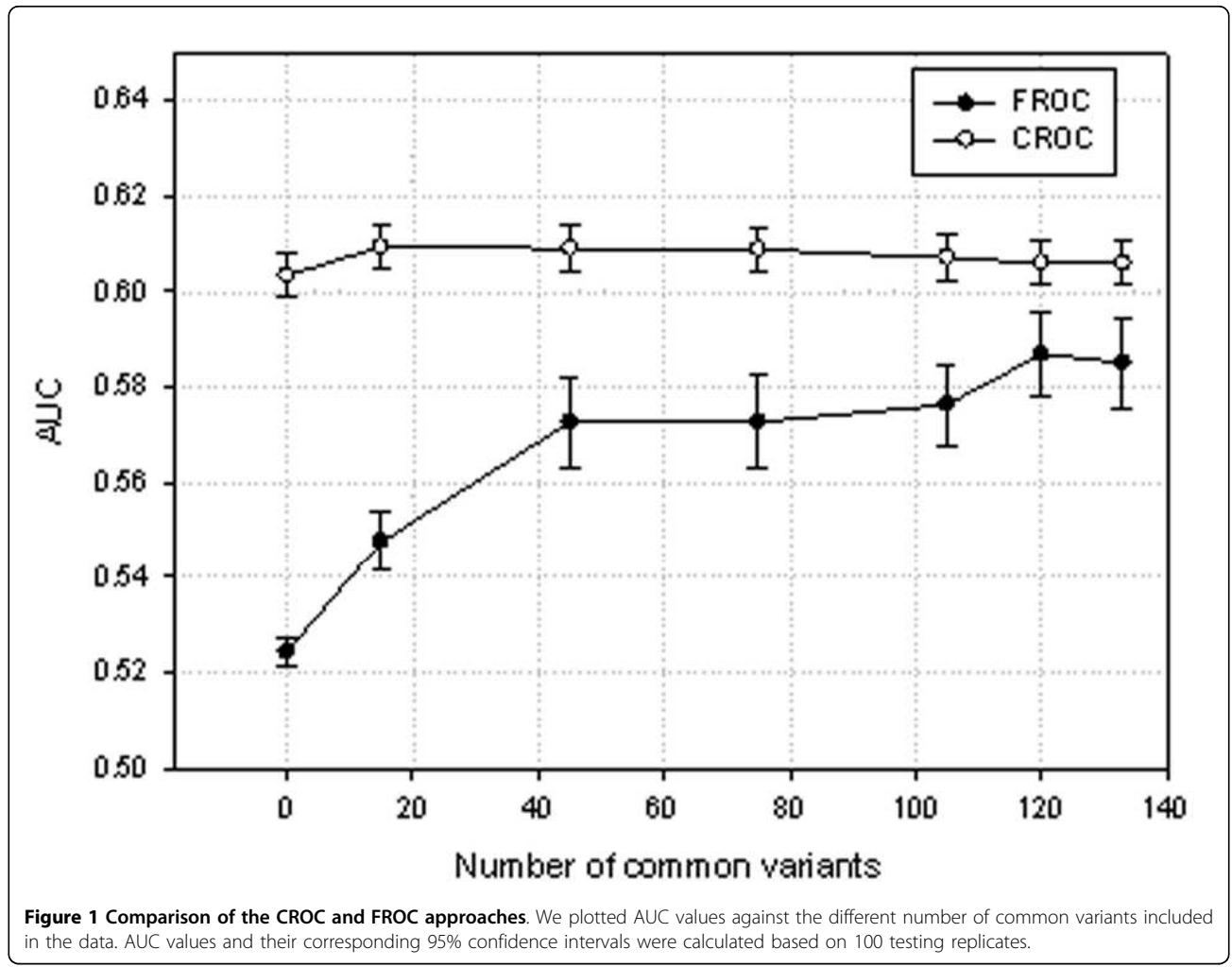

**Figure 1 Comparison of the CROC and FROC approaches**. We plotted AUC values against the different number of common variants included in the data. AUC values and their corresponding 95% confidence intervals were calculated based on 100 testing replicates.

of common variants. In the extreme scenario, in which only rare variants were used in the analysis, the CROC approach attained a much higher AUC value (0.603) than the FROC approach (0.524).

## Discussion

Next-generation sequencing technology is anticipated to be a powerful tool for uncovering novel genetic variants associated with complex disease, particularly for those variants with a low frequency. The rare variants identified through future whole-genome sequencing studies, if confirmed to have functional importance, may provide novel insights into underlying pathological and etiological processes. Yet, even if they are merely predictive and without functional importance, these rare variants can still be harnessed into clinical translational research applications. By incorporating these rare variants into current risk prediction models, we can predict disease outcomes more accurately.

We developed a CROC approach for future risk prediction research on sequencing data. The approach extends the previously developed FROC approach to deal with both common and rare variants by using a multistage collapsing procedure. The idea of collapsing was originally introduced in genetic association studies to deal with both common and rare variants. We have now integrated those ideas into the CROC approach for risk prediction research on sequencing data. By applying the approach on the simulated mini-exome sequencing data, we demonstrated the advantage of using both common and rare variants in risk prediction research. However, in this application, limited improvement was gained when additional rare variants were combined with common variants. This may be because rare variants account for only a small proportion of phenotype variation. However, in a different scenario, rare variants might contribute significantly to phenotype variation. Therefore we artificially decreased the influence of common variants. When the number of common variants was decreased, we found that significant improvement could be attained by considering additional rare variants. The accuracy of the risk prediction models can be further improved by considering environment risk predictors and gene-environment interactions. We ran an additional analysis including the environment risk predictors and found that the prediction accuracy was significantly improved (data not shown).

We considered only candidate genes in our analysis. Current risk prediction studies commonly adopt this strategy. However, for high-dimensional risk prediction research using millions of SNPs, variable selection becomes important. Although a forward selection algorithm is incorporated into the CROC approach, it could still be subject to false positives when dealing with whole-genome sequencing data. More sophisticated selection algorithms will be needed to deal with a large number of common and rare variants.

## Conclusion

We have developed a CROC approach for risk prediction analysis on sequencing data. By applying this new approach to the simulated GAW17 mini-exome sequencing data, we have illustrated that current risk prediction models built on common variants can be further improved by considering additional rare variants. In addition, we compared the CROC approach with the existing FROC approach. The CROC approach outperformed the FROC approach, especially when a large proportion of the considered variants were rare.

### Acknowledgments
This work was supported by start-up funds from Michigan State University. This article has been published as part of *BMC Proceedings* Volume 5 Supplement 9, 2011: Genetic Analysis Workshop 17. The full contents of the supplement are available online at http://www.biomedcentral.com/1753-6561/5?issue=S9.

### Authors' contributions
CW developed the statistical methods, conducted the analysis and drafted the manuscript. QL conceived of the study, developed the method, participated in its design and coordination and helped to draft the manuscript. Both authors read and approved the final manuscript.

### Competing interests
The authors declare that they have no competing interests.

Published: 29 November 2011